\documentclass[10pt,twocolumn,letterpaper]{article}

\usepackage[pagenumbers]{wacv} 

\usepackage{graphicx}
\usepackage{amsmath}
\usepackage{amssymb}
\usepackage{booktabs}
\usepackage{algpseudocode}
\usepackage{algorithm}
\usepackage{outlines}
\usepackage{bbm}
\usepackage{tabularx}
\usepackage{adjustbox}

\usepackage[pagebackref,breaklinks,colorlinks]{hyperref}

\usepackage[capitalize]{cleveref}
\crefname{section}{Sec.}{Secs.}
\Crefname{section}{Section}{Sections}
\Crefname{table}{Table}{Tables}
\crefname{table}{Tab.}{Tabs.}

\def\wacvPaperID{1516} 
\def\confName{WACV}
\def\confYear{2025}

\begin{document}

\title{Enhancing Worldwide Image Geolocation by Ensembling Satellite-Based Ground-Level Attribute Predictors}

\author{Michael J. Bianco \qquad\qquad\qquad
David Eigen \qquad\qquad\qquad
Michael Gormish\\
\qquad {\tt\small mike.bianco@clarifai.com} \qquad {\tt\small deigen@clarifai.com} \qquad
{\tt\small michael.gormish@clarifai.com}\\
Clarifai Inc., San Francisco
}
\maketitle

\begin{abstract}

We examine the challenge of estimating the location of a single ground-level image in the absence of GPS or other location metadata. These geolocation systems are currently evaluated by measuring the Great Circle Distance between a single predicted location and the ground truth.
Because this measurement only uses a single point, it cannot assess the quality of a set of regions or score heatmaps used by geolocation systems.  Accounting for larger areas in evaluation is required when there are follow-on procedures to further narrow down or verify the location.  This is especially important in rural areas, wilderness, and under-sampled regions, where finding the exact location may not be possible.


In this paper, we introduce a novel metric, Recall vs Area (RvA), which measures the accuracy of estimated distributions of locations.  RvA treats image geolocation results similarly to document retrieval, measuring recall as a function of area:  For a ranked list of (possibly non-contiguous) predicted regions, we measure the accumulated area required for the region to contain the ground truth coordinate.  This produces a curve similar to a precision-recall curve, where ``precision'' is replaced by square kilometers area, allowing evaluation of performance for different downstream search area budgets.

Following directly from this view of the problem, we then examine a simple ensembling approach to global-scale image geolocation, which incorporates information from multiple sources, and can readily incorporate multiple models, attribute predictors, and data sources.  We study its effectiveness by combining the geolocation models GeoEstimation \cite{muller2018geolocation} and the current state-of-the-art, GeoCLIP \cite{cepeda2023geoclip}, with attribute predictors based on Oak Ridge National Laboratory LandScan\cite{sims2022landscan} and European Space Agency Climate Change Initiative Land Cover \cite{esa2017}.  We find significant improvements in image geolocation for areas that are under-represented in the training set, particularly non-urban areas, on both Im2GPS3k and Street View images.

\end{abstract}

\vspace{-2mm}
\section{Introduction}
\label{sec:intro}


Accurate localization of ground-level imagery is important to a variety of applications, including navigation, tourism, and security, and has been the subject of recent research activity \cite{hays2008im2gps,weyand2016planet,muller2018geolocation,clark2023we,cepeda2023geoclip}.  Past works on this problem have primarily relied on publicly available geotagged image datasets to learn location estimators.
However, even large geotagged datasets
\cite{larson2017benchmarking,vo2017revisiting,muller2018geolocation,thomee2016yfcc100m,luo2022g} sample only a small fraction of the Earth's surface, concentrating mostly in cities and other population centers, with the most highly sampled locations corresponding to
the most highly trafficked sites, such as famous landmarks, parks, restaurants, etc.
Thus, when these geolocation systems are applied globally, many areas are under-sampled, and so applications in rural and other under-sampled areas are challenged by lack of representation in the training set.

To help address the challenge of generalization to under-sampled areas, 
we first introduce an novel image geolocation metric, {\it Recall vs Area} ({\it RvA}). This metric naturally characterizes model performance when predicted regions are discontiguous, which is important when the exact location may not be possible to recover, such as in applications to rural scenes or wilderness. Even though a single location may not be possible to predict in these situations, geolocation systems can still be used to suggest areas that can be studied by human analysts or further refined by additional automated systems.

Inspired by this view of the problem, we then examine a general ensembling approach to limit predicted area by incorporating ground-level attribute predictors corresponding to attributes visible by satellite. While ground level geotagged data are limited, satellite data have global coverage. Thus, accurate prediction of attributes mapped by satellite can extend to regions where there are few geotagged ground-level images.
For example, different land cover types (forest, mountains, roads, etc.) can be seen in both ground-level and satellite images. We consider two data products in our analysis: Oak Ridge National Laboratory (ORNL) LandScan Global \cite{sims2022landscan} and the European Space Agency Climate Change Initiative (ESA-CCI) Land Cover \cite{esa2017}.

We combine estimates from these attribute predictors with those from other global geolocation models, i.e. GeoEstimation\cite{muller2018geolocation} and GeoCLIP \cite{cepeda2023geoclip}, to improve performance on rural regions under-sampled in the geotagged data.  We evaluate our method using both the standard Im2GPS3k \cite{vo2017revisiting}, as well as a Street View dataset (see Sec.~\ref{sec:datasets}) that is different in geographic distribution and scene concepts.

\begin{figure*}
  \centering
  \includegraphics[width=1.\textwidth]{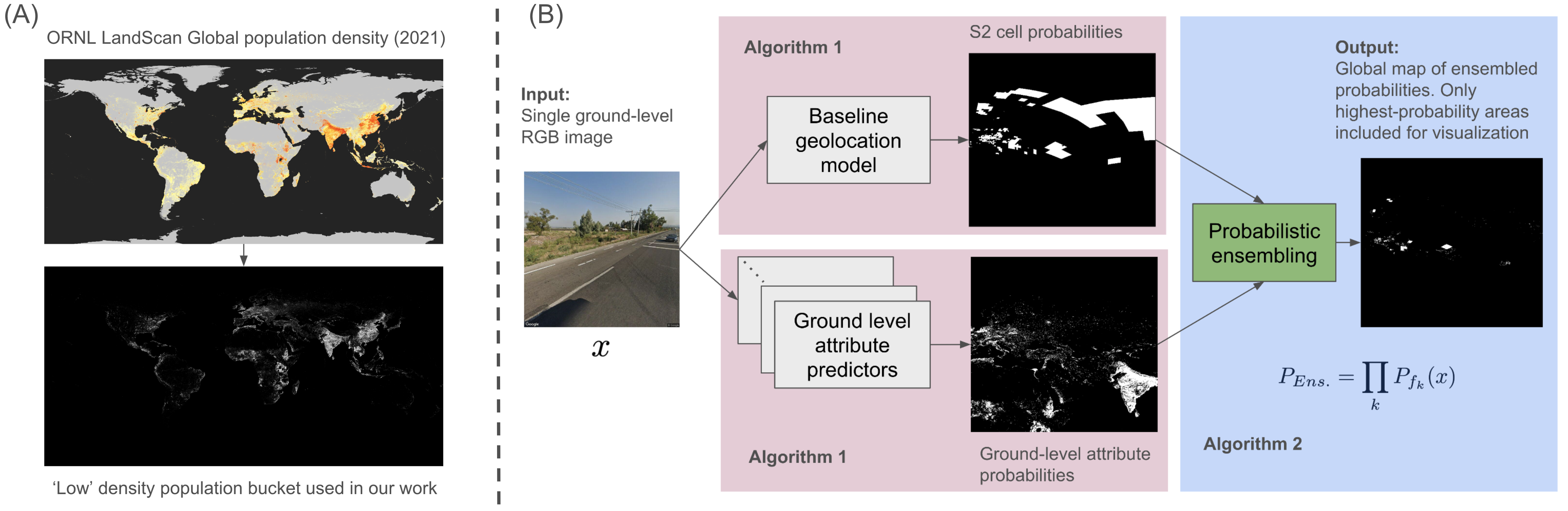}
  \caption{(A) Visualization of `low' population density mask obtained from ORNL LandScan global, see Sec.~\ref{sec:predictors} for discussion. Masks obtained from LandScan as well as ESA-CCI Land Cover are used by our ensembling method to improve image location estimates of base geolocation models. (B) The objective of our proposed ensembling approach is, given a single ground-level RGB image, to obtain a global geolocation probability map that incorporates information from base geolocation models (in this work, GeoEstimation and GeoCLIP) and ground-level maps obtained from satellite data products. The image is passed to both the base geolocation model and the individual ground-level attribute predictors. Example output $S^2$ cell and ground level attribute probabilities from Algo.~\ref{algo:geoest} and ~\ref{algo:model_isec}, and final ensembled probabilities are shown. The high probability area is significantly reduced in ensembling. Probability maps thresholded for visualization.}\label{fig:framework}
\end{figure*}

\begin{figure}
  \centering
  \begin{subfigure}[t]{0.23\textwidth}
    \includegraphics[width=\textwidth]{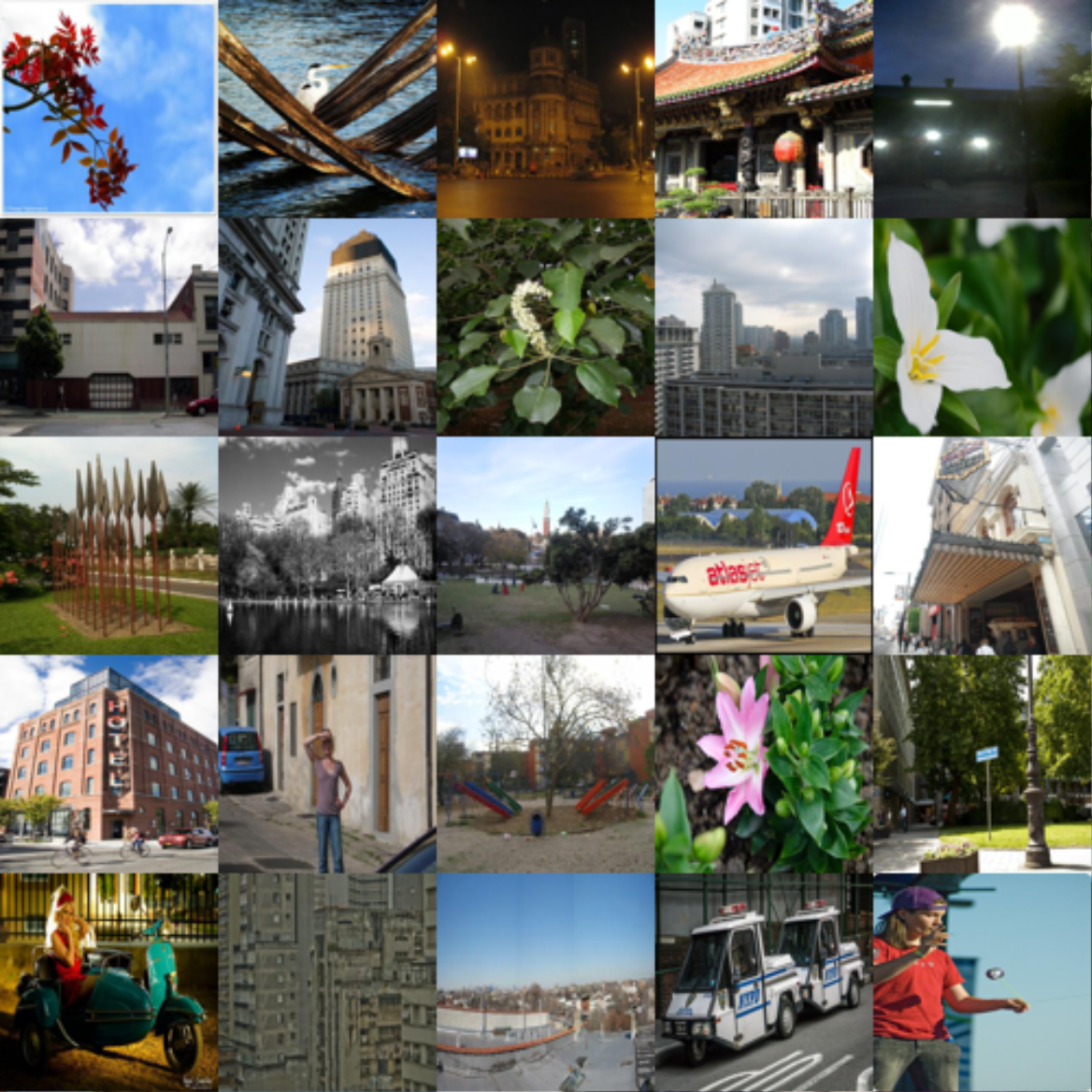}
  \end{subfigure} 
  \begin{subfigure}[t]{0.23\textwidth}
    \includegraphics[width=\textwidth]{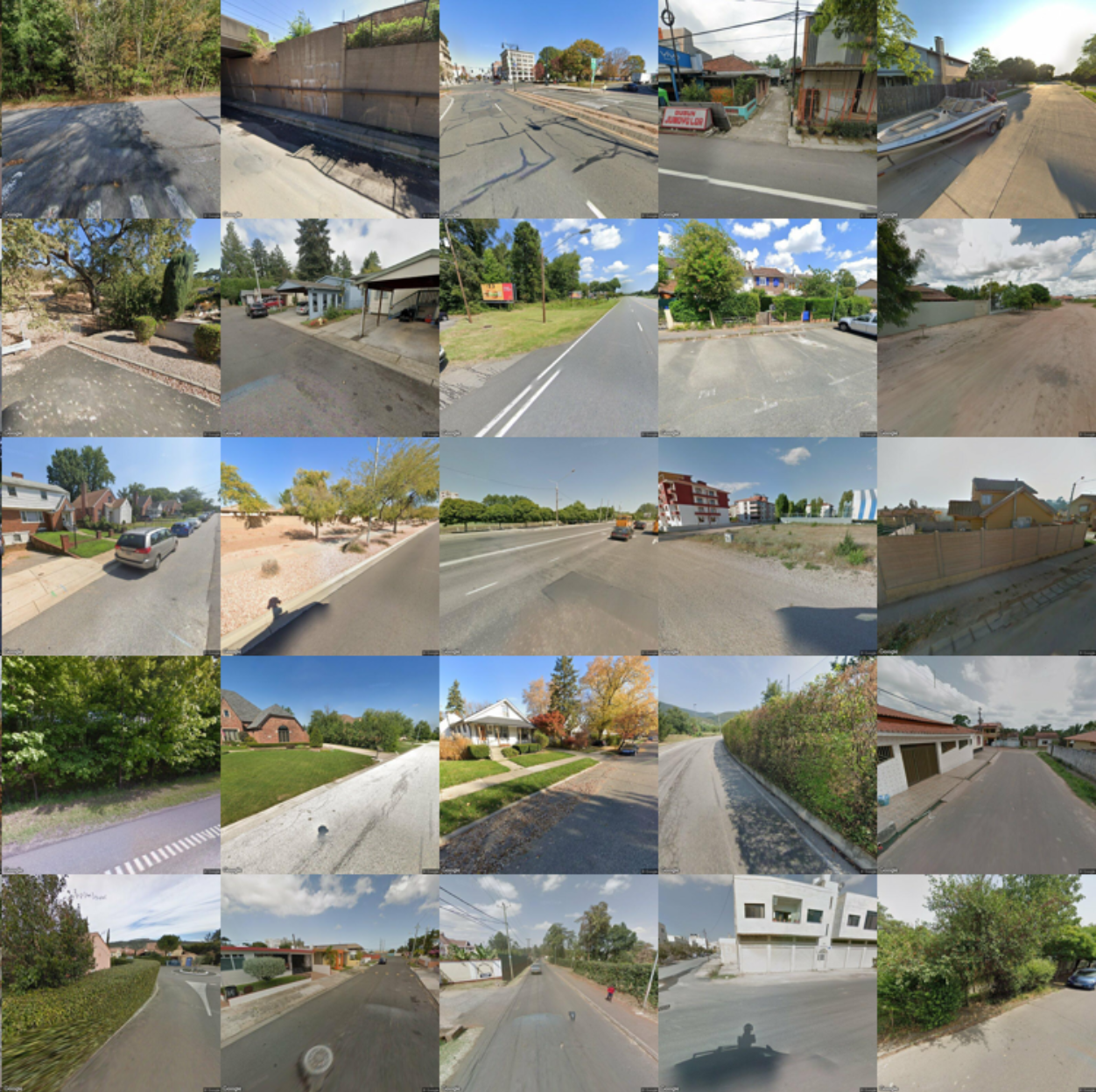}
  \end{subfigure}
  \caption{Batches of images from (left) MP-16 and (right) Street View, from medium urban density regions per LandScan labeling (see Sec.~\ref{sec:datasets}, \ref{sec:predictors}). In general, the MP-16 dataset consists of more typical urban scenes, whereas Street View contains more scenery from near roads. Further MP-16 dataset is drawn from areas much more close to urban environments.}
  \label{fig:mp16_streetview_images}
  \vspace{-3mm}
\end{figure}

\subsection{Contributions}
\begin{itemize}
    \item We develop a novel metric, Recall vs Area (RvA), for evaluating geolocation systems, which can account for predictions of multiple, non-contiguous areas. We use this method to assess image geolocation approaches.
    
    \item We show that, in isolation, geolocation methods trained only on geotagged imagery can be vulnerable to image domain shift with respect to training data.

    \item We develop an approach to combine
    ground-level attribute predictors with geolocation models, leveraging satellite data products, and show that incorporating the satellite information improves generalization to areas of the Earth for which there are few ground-level picture samples.
    
    
    
\end{itemize}

\section{Related Work}

M{\"u}ller-Budack et al.\cite{muller2018geolocation} develop a hierarchical cell-based approach to geolocation. They fine-tune their model on the large MP-16 dataset\cite{larson2017benchmarking}, and quantize the Earth using $S^2$ cells (using the Google $S2$ geometry Library\footnote{\url{https://code.google.com/archive/p/s2-geometry-library/}}), with three resolutions $\{\text{`coarse',`medium',`fine'}\}$ depending on data density. Their model outputs image geolocations for each level, whose scores combined to create a final estimate over the finest level cells. 
Subsequent papers have built on this work, proposing various cell division strategies, including CPlaNet \cite{seo2018cplanet} and Translocator \cite{pramanick2022world}.  These approaches rely exclusively on geotagged ground-level images as their datasource, which are assembled into cell bins of different types.

Other recent works have brought additional information to the geolocation problem.  
Lin et al.\cite{lin2013cross} explored using land-cover information for improving query-based geolocation, but required associated triplets of ground-level and aerial imagery and land use and was applied only to a small 40$\times$40~km region.  More recent works include relationships between the scene and variations in environment\cite{clark2023we} or from segmentation models\cite{pramanick2022world}. Luo et al.\cite{luo2022g} and Cepeda et al.\cite{cepeda2023geoclip} exploit geographical correlations encoded in vision-language foundation models (using Contrastive Language-Image Pretraining\cite{radford2021learning}) to improve image geolocation performance.

There is also significant work in cross-view image localization, e.g. \cite{lin2013cross, wang2024fine, hu2018cvm}. In these approaches, optical satellite imagery is matched to ground-level imagery to achieve ground-level image localization. These methods are used to pinpoint exact locations within limited-area urban scenes where dense ground-level imagery is available. This is complementary to our work, which uses ground-level attributes and satellite sources to limit search areas over a global field; indeed, these works are examples of systems that might run downstream of ours within a proposed area.

As far as we are aware, the approach we present is the first to use land cover and other attributes recognizable from ground level for global-scale geolocation.
We combine predictions from multiple sources, attribute predictors associated with satellite-based maps, along with cell-based estimation systems, using a common ensembling grid.
This yields an effective intersection of non-contiguous regions predicted from the available evidence to limit the geolocation area.

\section{Approach}
\subsection{Recall vs Area Metric}

Existing approaches typically measure geolocalization accuracy by calculating the distance of a single predicted location from the ground truth.  Distance is calculated using great circle distance (GCD), and binned into a smaller handful of thresholds.  The fraction of images whose top prediction lies within each threshold is reported as the accuracy for that distance. For example \cite{muller2018geolocation} used GCD thresholds of $[1, 25, 200, 750,2500]$~km to evaluate model performance, even giving these distances names to represent how close the solution was, namely street, city, region, country, continent.

While this measure works well to characterize performance of a single predicted location, it does not extend to measuring performance of a larger (possibly discontiguous) predicted area.  This is important for many practical applications, where the image geolocation algorithm is used to propose a set of possible locations that are used in downstream tasks. Thus, we would like to measure localization performance in terms of an area budget.

In addition, many distant regions on Earth look similar from ground-level, as illustrated by the often scattered global distribution of high confidence $S^2$ cell predictions by GeoEstimation (see e.g. Fig.~\ref{fig:framework}(b)).  In the context of a system that progressively narrows down the possible locations, it is important to maintain recall at each stage while still reducing the set of predicted regions.


We propose a new metric, Recall vs Area ({\it RvA}), to characterize image geolocation model performance. We rely on the conventions of recall and precision, and formulate a geographic analogy to document retrieval. For each image, a model predicts scores over a set of region proposal areas. To evaluate, we accumulate areas by score, up to a given search budget. Recall is determined according to whether the ground truth lies in the accumulated area. We then compute recall for different area budgets (thresholds), resulting in a ROC-like curve comparing recall on one axis and area budgets on the other.

Here, recall is determined by whether the geographic area returned by the model for a query image contains the image's location.  More formally, for a query image $x_i\in\mathcal{X}$ with ground truth location $y_i$, a geolocation model will return a list of location areas ($S^2$ cells, points, etc.) $a_{ik}, k \in [1..K]$, along with confidence scores.  We accumulate area sorted by model confidence, to create a (possibly discontiguous) region, up to an area budget (threshold).  For an area threshold $\alpha$, recall is then
\begin{align} \label{eq:recall_fcn_area}
    \text{\it recall}(\alpha) = 
    \frac
        {\sum_i \mathbbm{1}[
            y_i \in \{a_{ik} : \sum_{k'=1}^{k} a_{ik'} \leq\alpha\}]}
        {|\mathcal{X}|},
\end{align}
{\it i.e.} the fraction of images whose ground-truth location lies within the model's top-scoring predicted areas, up to the area threshold $\alpha$.

%


We then plot the curve $recall(\alpha)$ for different values of the area budget $\alpha$, resulting in a description of model performance, akin to a precision-recall curve (see Figs.~\ref{fig:geoest_recall_vs_area} and \ref{fig:geoclip_recall_vs_area}).\footnote{Indeed, area itself can be thought of as a notion of inverse precision, with smaller areas being more precise and larger ones including more false positive locations and therefore less precise.}  The curve is used to evaluate model performance, and can also be used when selecting appropriate trade-offs between recall and size of a proposed area.

\subsection{Algorithm} \label{sec:ensembling_algo}


Given an image $x$, we ensemble geolocation and ground-level attribute predictors to find a subset of areas on Earth most likely to contain the image. Fig.~\ref{fig:framework} illustrates this process.


Because we combine the geolocation area predictions of multiple models, we map each model's output to a common grid over the surface of the Earth, with ``pixel" coordinates given by latitude and longitude (WGS-84 projection). Note that we index by (lat, lon), so each ``pixel'' has a different area on the Earth depending on its latitude coordinate. To account for this, we also create an array $A_{pix}$ that contains the surface area corresponding to each point in the map. 

The output of each model $f(\cdot)$ is a vector of probabilities $p=[p_{1}\dots p_{J}]\in\mathbb{R}^J$, corresponding a set of binary masks $\mathcal{M}_f=\{m_{1}\dots m_{J}\}$, with $|\mathcal{M}|=J$, and $m\in\{0,1\}^{H\times W}$ with $H$ and $W$ the vertical (latitude) and horizontal (longitude) resolution of the entire surface grid. For GeoEstimation \cite{muller2018geolocation}, each mask is an $S^2$ cell and each probability is the model's softmax score for the corresponding cell. To map GeoCLIP\cite{cepeda2023geoclip} predictions to the common grid, we use the same $S^2$ cells as in GeoEstimation, since GeoCLIP was trained using MP-16, which is the same dataset used to distribute the $S^2$ cells.  We obtain a score for each cell by passing the center location to GeoCLIP.

For LandScan and LandCover attribute predictors, each mask $m_j$ is a discontiguous region corresponding to a population density bucket, land use type, etc., and probabilities are the softmax scores for each attribute. 

The scores for the attribute masks and $S^2$ cells are assigned to the common grid, normalizing by their areas.





\begin{algorithm}[t!]
\caption{Assigning probabilities to the common grid}\label{algo:geoest}
\begin{algorithmic}
\State Given image $x$, model $f$, and mask(s) $\mathcal M_f$
\State  {\it \# compute softmax score for each mask using the model}
\State $p = f(x)$
\State {\it \# combine into rasterized grid}
\State $P_f = \sum_{j=1} {p_{j}m_{j}}/{area(m_{j})}$, with $m_j\in\mathcal{M}_f$
\State \Return ${P_f}$
\end{algorithmic}
\end{algorithm}
\begin{algorithm}[t!]
\caption{Ensembling probabilities from ground-level attribute predictors and  GeoEst./GeoCLIP}\label{algo:model_isec}
\begin{algorithmic}
\State Given $x, \{{f}_1\dots{f}_K\},\{\mathcal{M}_1\dots\mathcal{M}_K\} $
\State \Return $P = \prod_{k=1}^K  P_{f_k}$, with $P_{f_k}$ as in Alg.1
\end{algorithmic}
\end{algorithm}
\begin{algorithm}[t!]
\caption{Model evaluation}\label{algo:eval}
\begin{algorithmic}
\State Given dataset $\mathcal{D},\ \mathrm{model}$, pixel spherical areas map $A_{pix}$
\State Returns accumulated areas necessary to the include ground truth location for each image.
\State $\mathcal{A}\gets\{\}$ {\it \# eval. areas}
\For {$(\mathrm{image}~x_i, \mathrm{g.t. location}~y_i) \in \mathcal{D}$}
\State $P\gets \ \mathrm{model}(x)$
\State $p^* = P[y_i]$ {\it \# predicted prob at ground truth loc.}
\State $area = \sum A_{pix} \cdot \mathbbm{1}[P \ge p^*]$  {\it \# area of pred $\ge p^*$}
\State $\mathcal{A}\gets \mathcal{A}\cup area$
\EndFor
\State \Return $\mathcal{A}$
\end{algorithmic}
\end{algorithm}

We then combine probabilities to produce a final set of scores, estimated over the common grid.  For this, we use a simple element-wise product of probability maps, which we found to be effective even though the masks are not independent:
\begin{align} \label{eq:independent_attributes}
    P_{Ens.} = \prod_k P_{f_k}(x)
\end{align}
where each $P_{f_k}(x)$ is the probability map of the $k^\mathrm{th}$ predictor (GeoEst, Land Cover regions, etc.) as described in Algorithms \ref{algo:geoest} and \ref{algo:model_isec}.
For evaluation, locations (pixels) on the common grid are accumulated in score order for each image until the resulting region contains the ground truth location.  Note that since we index the grid by (lat, lon), each element has a different area according to latitude, we accumulate area using a pre-computed matrix $A_{pix}$ that contains the area of each pixel.  This process is defined in Algorithm~\ref{algo:eval}.  This gives us an array with the smallest cumulative area needed to contain the ground truth for each image --- i.e. the smallest area threshold $\alpha$ at which each image prediction is a ``hit''.  We then sort this array to produce the RvA curve. In deployment, the area threshold can be used to calibrate between desired recall and an area budget.

\section{Datasets}

\subsection{Ground-level attribute datasets and processing}


In this work, we utilize two satellite data products: LandScan Global \cite{sims2022landscan} from ORNL and Land Cover \cite{esa2017}, from the ESA-CCI. Both data products provide high-resolution estimates of surface attributes: LandScan Global predicts 24-hour average population density, while ESA Land Cover classifies 38 natural and anthropogenic land cover types.

We used three image datasets. MP-16 \cite{thomee2016yfcc100m} and Im2GPS3k \cite{vo2017revisiting} are commonly used training and evaluation benchmarks in image geolocation research. MP-16 was employed to train our attribute classifiers, and Im2GPS3k was used for testing. Additionally, we collected images from Google Street View, using training-set panorama IDs from \cite{luo2022g}, to assess how well the models generalize to image location distributions and capture pipelines that differ from MP-16 and Im2GPS3k.


\begin{figure}
  \centering\adjincludegraphics[trim={0.25\width} {0.25\height} {0.25\width} {0.05\height} ,clip,width=0.95\columnwidth]{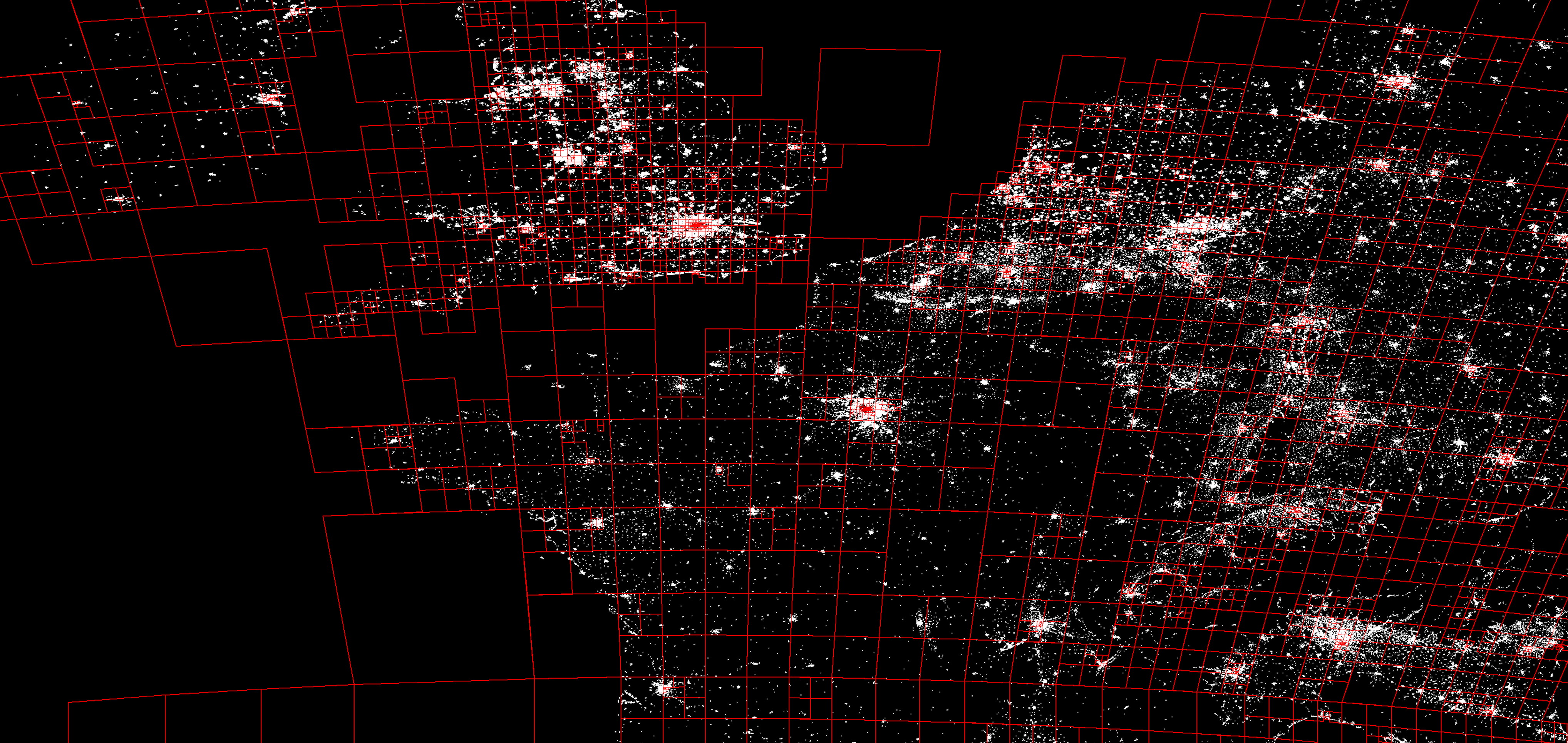}
    \caption{Example overlay centered on Europe, showing $S^2$ cells (red boxes, fine level from GeoEstimation \cite{muller2018geolocation}) with LandScan\cite{sims2022landscan} medium population density mask (white). The mask is much more restrictive than the $S^2$ cells alone, yielding improvements in RvA from model ensembling.}
    \label{fig:s2_landscan_overlay}
\end{figure}

\subsubsection{LandScan Global}
The ORNL LandScan Global dataset provides a global map of population density at a resolution of approximately $1~\text{km}^2$ \cite{sims2022landscan}, as shown in  Fig.~\ref{fig:framework}(a). The data is a large TIFF image, $21600\times 43200$ pixels (lat x lon), and contains 24-hour average population densities. 

Each pixel indicates whether the corresponding region is land and if so, its value contains the population density. The image uses WGS-84 coordinates, a standard lat-lon representation, with a resolution of approximately $1~\text{km}^2$.


\begin{figure*}[ht]
  \centering
  \setlength{\tabcolsep}{0mm}
  \renewcommand{\arraystretch}{0}

    \begin{tabular}{ccc}
    
    {\fontfamily{cmss}\selectfont \qquad RvA} &
    {\fontfamily{cmss}\selectfont \quad RvA (Rebalanced)} &
    {\fontfamily{cmss}\selectfont \quad Absolute Improvement (Rebalanced)} 
    \\
    
    \includegraphics[width=0.7\columnwidth]{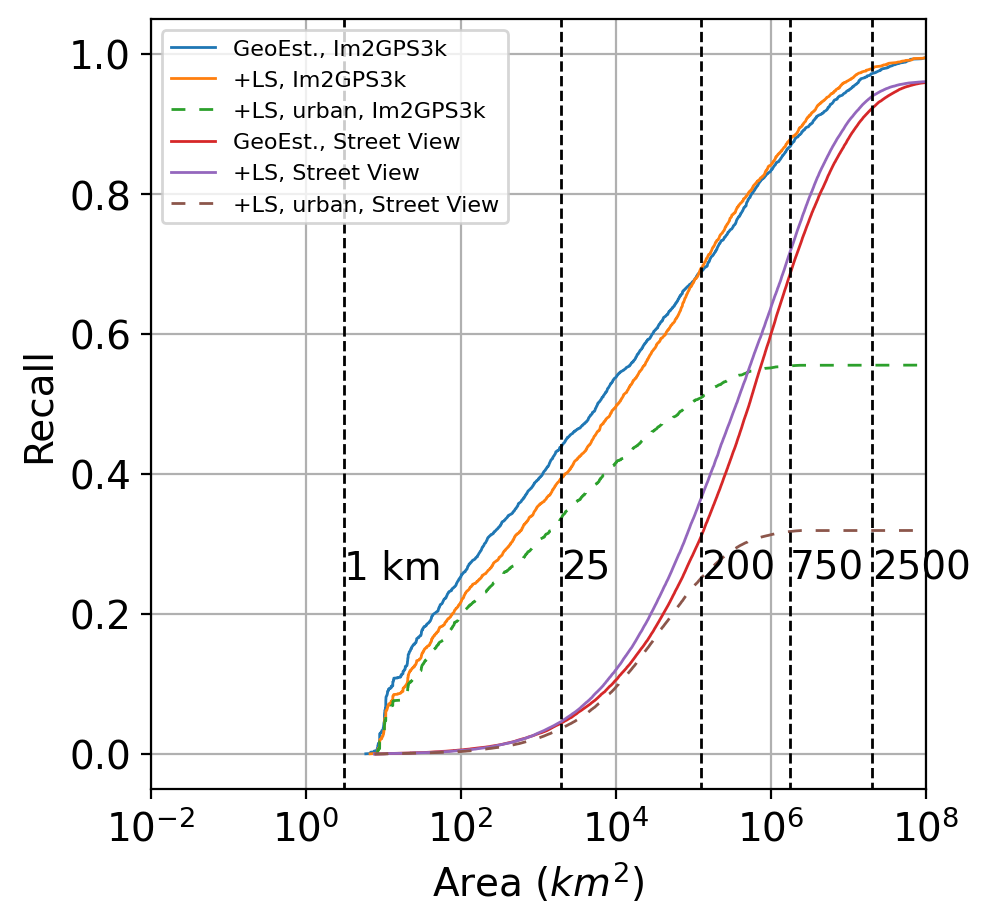}
    &
    \includegraphics[width=0.7\columnwidth]{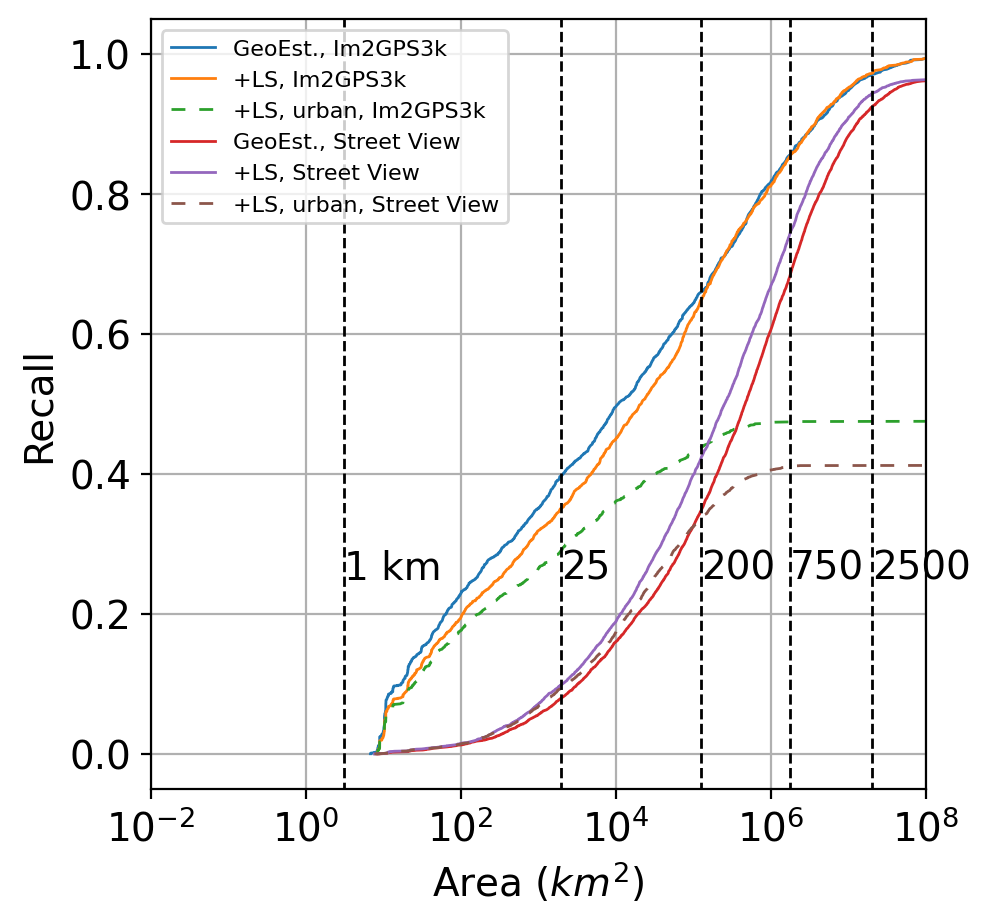}

    &
    \includegraphics[width=0.7\columnwidth]{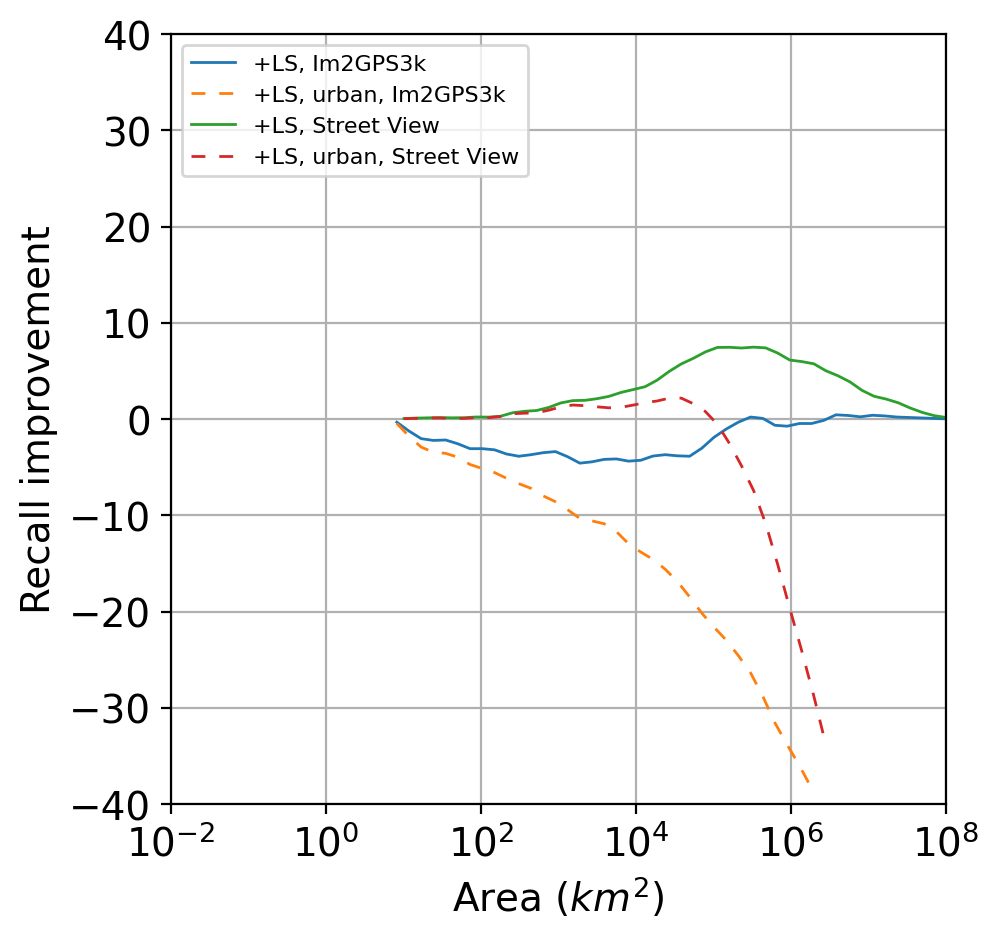}
    \\

    (a) & (b) & (c) \\

    \end{tabular}

  \caption{(a) {\it Recall vs. Area} ({\it RvA}) obtained for
  rasterized GeoEst (Alg.\ref{algo:geoest}), alone and ensembled with LandScan (+LS) attribute prediction.  As an additional baseline, we compare always applying the LS ``urban'' mask instead of predicting the bucket (+urban, dotted curve). Spherical cap areas shown in blacked dashed lines, see Sec.~\ref{sec:results} for details.  (b) RvA, measured on data rebalanced over urban and non-urban areas by randomly sampling equal number of images from each LS mask. Our method improves performance for less-populated areas while maintaining urban area results. (c) Relative improvement measured on the balanced datasets.}\label{fig:geoest_recall_vs_area}
\end{figure*}

\begin{figure*}[ht]
  \centering
  \setlength{\tabcolsep}{0mm}
  \renewcommand{\arraystretch}{0}

    \begin{tabular}{ccc}
    
    {\fontfamily{cmss}\selectfont \qquad RvA} &
    {\fontfamily{cmss}\selectfont \quad RvA Area (Rebalanced)} &
    {\fontfamily{cmss}\selectfont \quad Absolute Improvement (Rebalanced)} 
    \\
    
    \includegraphics[width=0.7\columnwidth]{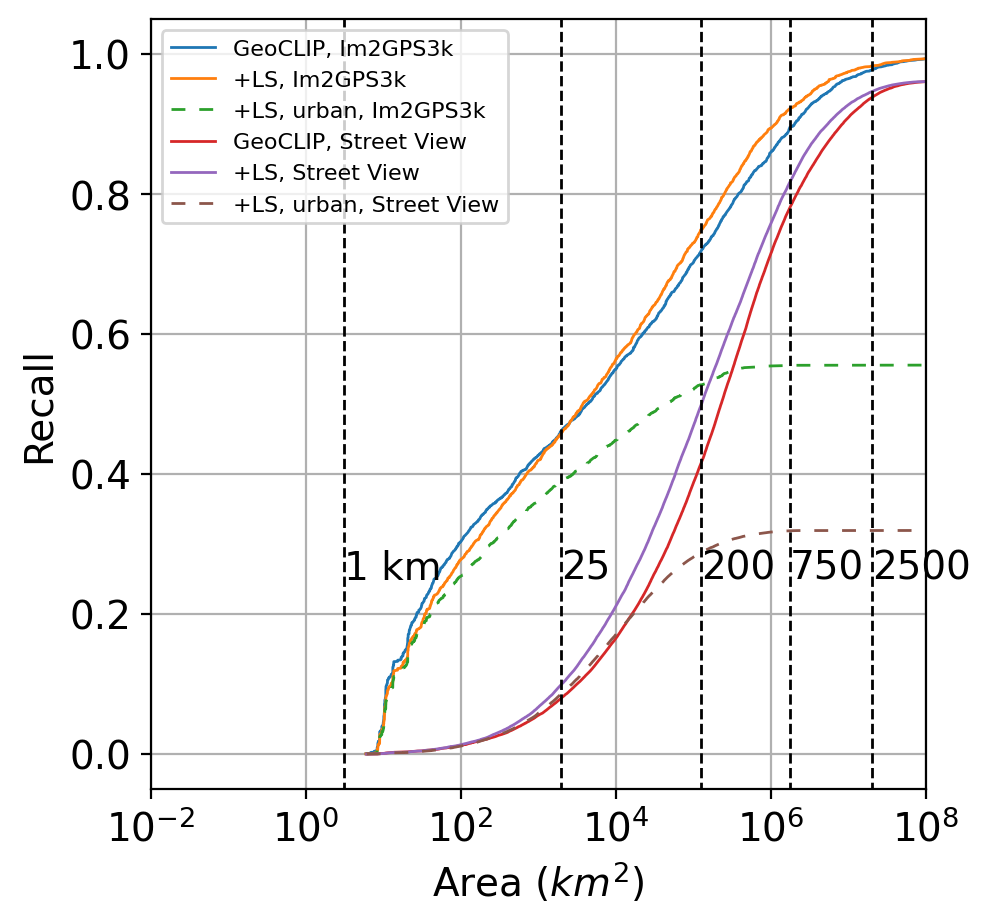}
    &
    \includegraphics[width=0.7\columnwidth]{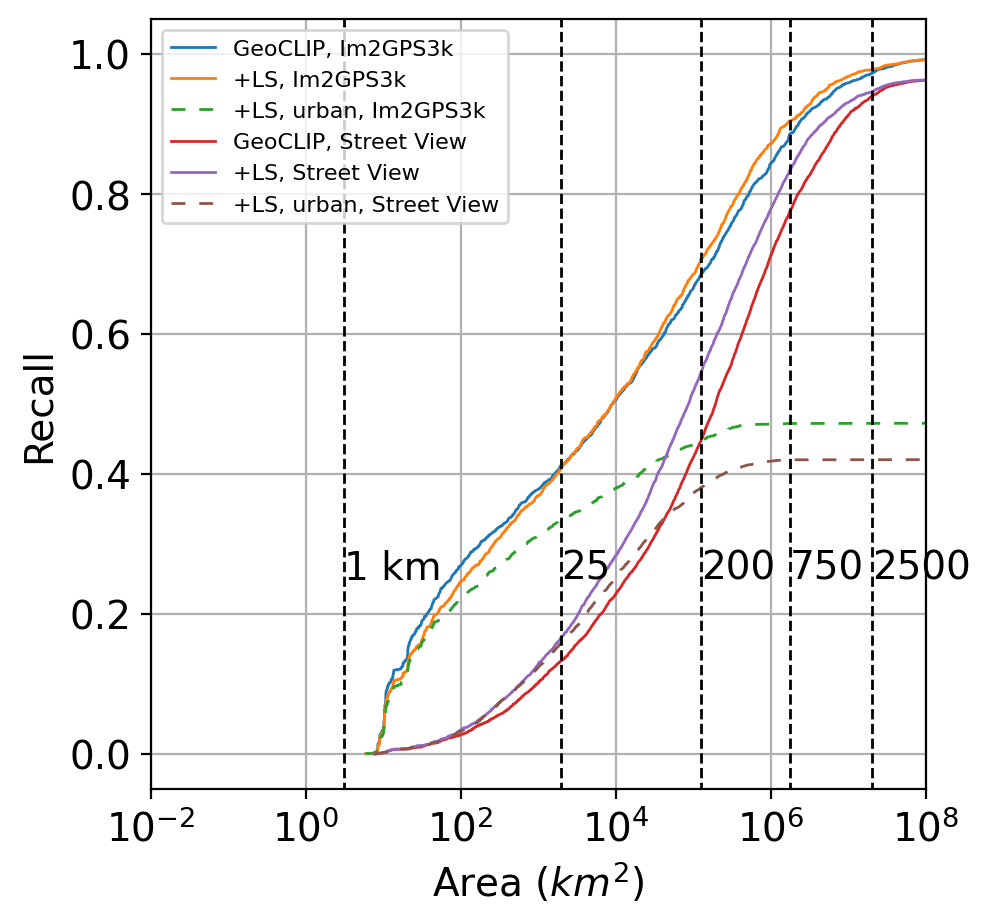}
   
    &
    \includegraphics[width=0.7\columnwidth]{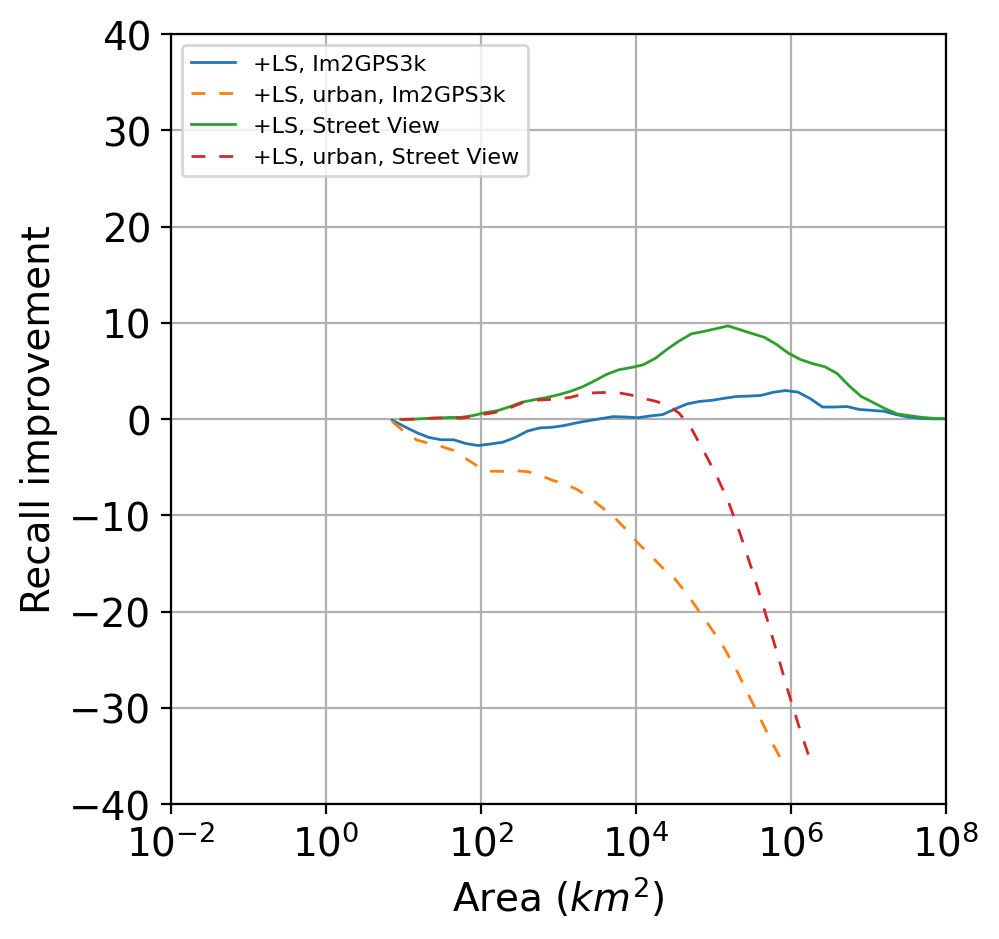}
    \\

    (a) & (b) & (c) \\

    \end{tabular}
  \vspace{-2mm}

  \caption{Rasterized GeoCLIP \ref{algo:geoest} and ensembling: same analysis as Fig. \ref{fig:geoest_recall_vs_area}(a--c).} \label{fig:geoclip_recall_vs_area}
  \vspace{-2mm}
\end{figure*}

\subsubsection{ESA Landcover}
ESA-CCI Land Cover is a global land cover and land use data product \cite{esa2017}. The attributes are predicted using deep learning classifiers applied to Sentinel-2 satellite images. The dataset is a $64800\times 129600$ pixel (lat x lon, WGS-84) GeoTIFF with 300 m resolution. Each pixel is labeled with the most likely land cover from 38 classes (grouped into 22 super-classes), including cropland, grassland, tree cover, and developed land use categories. Maps are available for the years 1992-2020. We used the Land Cover 2015 dataset for the development of region reduction classifiers.




\subsection{Image datasets} \label{sec:datasets}
The most comprehensive geotagged ground-level image dataset currently available is MP-16 \cite{larson2017benchmarking}, a subset of the Yahoo Flickr Creative Commons 100 Million datasets (YFCC100M)\cite{thomee2016yfcc100m} that includes geotags. This dataset was used to train ground-level attribute predictors.

The size of the dataset is ~4.7M images.  of which we were able to obtain 4.2 million, as approximately 500~k images were no longer available at the URLs provided by \cite{muller2018geolocation}. 
The dataset includes a diverse range of image concepts and objects, featuring both indoor and outdoor scenes of natural and human-made environments.

MP-16 was originally the training set used in the GeoEstimation paper \cite{muller2018geolocation} and continues to be used in recent research, such as in \cite{cepeda2023geoclip,pramanick2022world}. In addition, we use the Im2GPS3k dataset \cite{vo2017revisiting} as a benchmark for testing our models.

For evaluation, we rely on Im2GPS3k \cite{vo2017revisiting} and a set of images from Google Street View \cite{streetview}. To specifically assess the models' generalization and robustness to domain gaps, we do not train on any Street View images and use them exclusively for evaluation.


The Google Street View data \cite{streetview} was obtained using panorama IDs from Luo et al. \cite{luo2022g}. This geo-diverse dataset includes at least 426 panoramas from 90 countries. From this overall distribution, we randomly constructed train, validation, and test sets. We first sampled 50000 panorama IDs from a total of 322536 panoramas. Of the 50000 IDs chosen, 49829 IDs were still valid. We thus obtained 49829 images---each a single 90° FOV slice from a full panorama, with 15° downward pitch (toward ground to capture road markings) from the Street View API, along with corresponding image geotags.

\section{Experiments}

\subsection{Ground-level attribute predictors}\label{sec:predictors}

We trained two ground-level attribute predictors, one for each satellite data product (LandScan and Land Cover), using MP-16 imagery relabeled with attributes from each source. In our experiments, we use the acronyms LS for LandScan and LC for Land Cover. For each image, the ground-level attribute labels were assigned by indexing the nearest pixel value in the attribute maps at the native TIFF resolution. As described in Section~\ref{sec:ensembling_algo} the outputs of these models are projected on global masks and combined to find a global distribution of probable locations.


  
\begin{figure}[t!]
\centering
\begin{subfigure}{0.25\textwidth}
 \includegraphics[width=\textwidth]{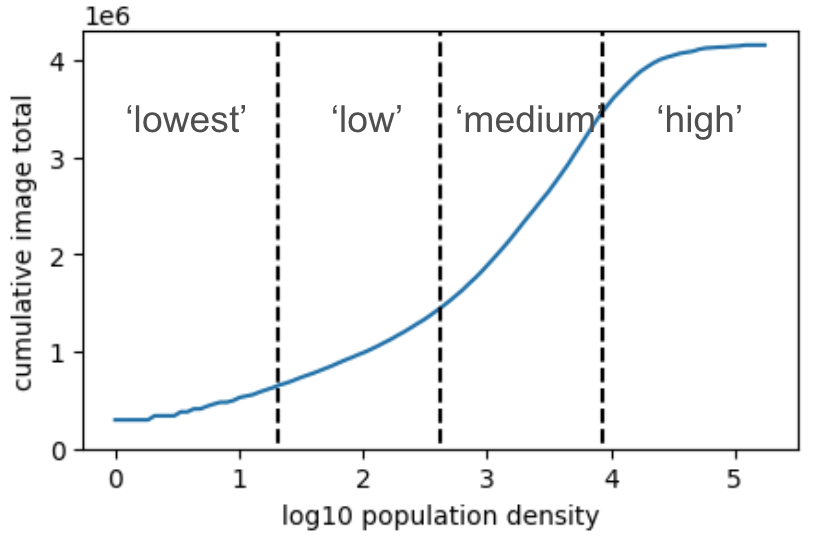}
  \caption{} 
\end{subfigure}~~
\begin{subfigure}{0.2\textwidth}
 \includegraphics[width=\textwidth]{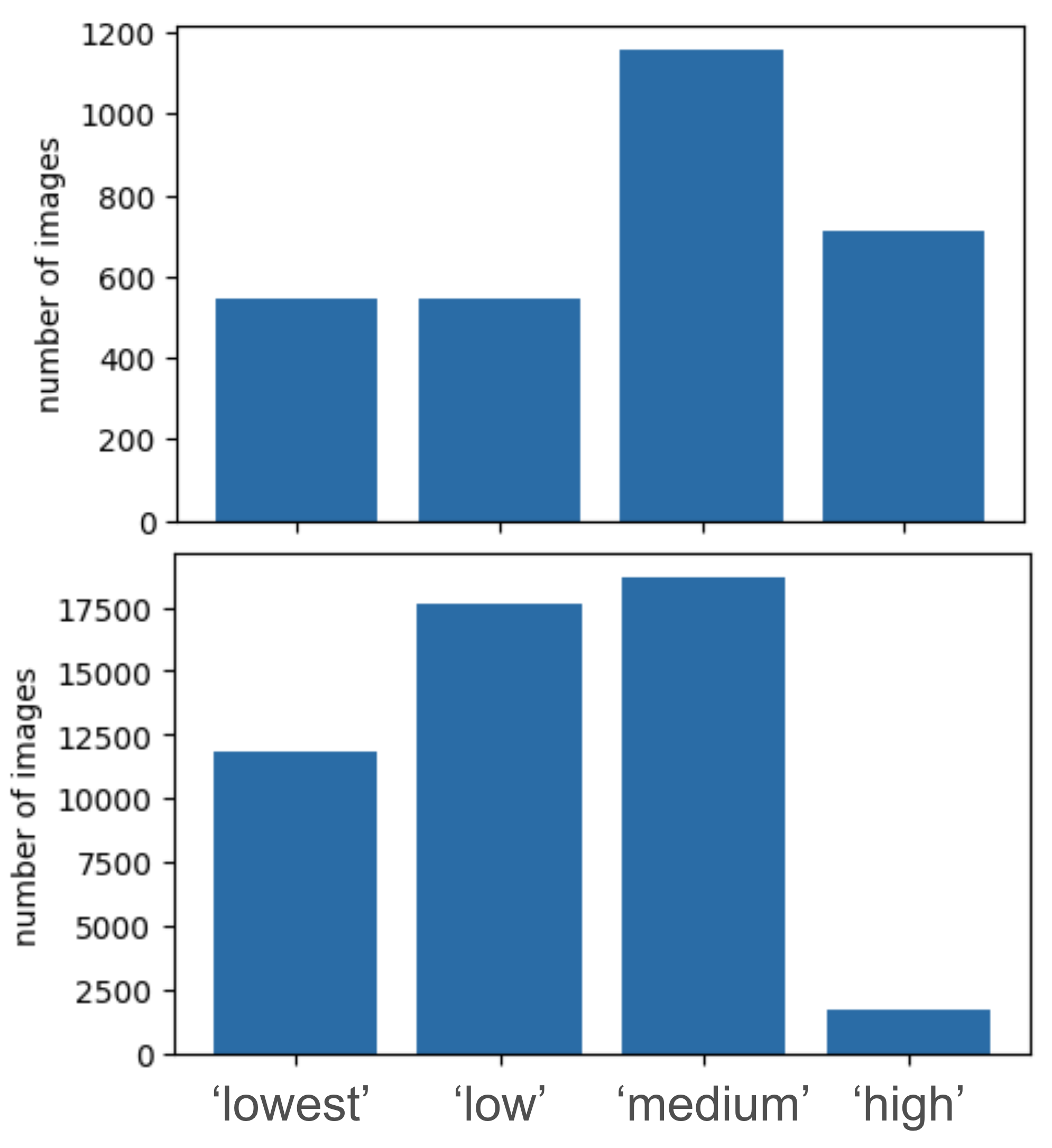}
  \caption{\label{fig:streetview_vs_im2gps3k_dist}}
\end{subfigure}
    \caption{(a) Cumulative distribution of MP-16 LandScan (LS) population density. LS buckets shown as $[\text{`lowest'}, \text{`low'}, \text{`medium'}, \text{`high'}]$.  (b) Distribution of original (unbalanced) Im2GPS3k (top) and Street View images (bottom) in LandScan (LS) buckets.}\label{fig:ls_buckets}
    \vspace{-2mm}
\end{figure}

Since the LS population densities are continuous, we categorized these into four density buckets for use in a softmax classifier. These buckets were created based on the $\log_{10}$ of population density, using equal bin widths over the range of LS population densities.  Fig.~\ref{fig:ls_buckets} shows the LS population density range sampled by MP-16 along with the bucket boundaries.
This results in four buckets of
$[\text{`lowest'}, \text{`low'}, \text{`medium'}, \text{`high'}]$ density, cover land areas in percentages $[93.9, 4.93, 1.10, 0.029]$\%, respectively.  Thus the LS classifier predicts these 4 classes. In Fig.~\ref{fig:s2_landscan_overlay} the medium population density bucket is overlaid with the $S^2$ cells (fine) used for $\text{base}\big(M,f^*\big)$. The population density mask is more restrictive than the $S^2$ cells alone.

For LC, we labeled the MP-16 images using the 22 top-level land cover classes and then merged these into 7 broader classes that our classifier predicts (see Sec. \ref{sec:lc-merge}).

To balance the classes during training, we limited the images in the 'urban area' LC category to $12.5~\%$ of the MP-16 dataset. This adjustment was necessary because most MP-16 images are from densely populated areas, while our goal is to locate both urban and rural locations.

For our predictors, we used a pretrained ResNet-50v2 architecture from \cite{kolesnikov2020big}, as implemented in the Pytorch Image Models (timm) library \cite{rw2019timm}. 
Specifically, we use the Big Transfer (BIT)-M version of the model, which is pretrained on the public ImageNet21k dataset \cite{ridnik2021imagenet21k}.




The predictors were trained on the relabeled MP-16 imagery for each of the LS classes for a maximum of 100 epochs using distributed training over 4 GPUs. In training, the images were augmented using random cropping, rescaling, horizontal and vertical flipping, and color jitter. 

\subsection{Ensembling} \label{sec:ensembling}

In our experiments, the common grid size was $H=5400$ (latitude) by $W=10800$ (longitude). These dimensions were chosen because they are common factors of the LS and LC TIFF image sizes, and are small enough to yield reasonable time complexity and geolocalization performance. The average resolution of the common image is $\sim4~\text{km}$, which corresponds to zero recall at $\sim16~\text{km}^2$ in Figs.~\ref{fig:geoest_recall_vs_area} and \ref{fig:geoclip_recall_vs_area},  and Tables~\ref{table:results_im2gps} and \ref{table:results_streetview}. 

For GeoEstimation, we used the published hierarchical model (``$\text{base}(M,f^*)$'' from \cite{muller2018geolocation}). For GeoCLIP, we used the published model architecture and weights. Because both GeoEst. and GeoCLIP are fine-tuned with MP-16, we use the $S^2$ cell centers, which are also distributed according to MP-16, to sample the GeoCLIP location encoder and obtain scores for each of the S2 cells.

To generate the probability map $p\in\mathbb{R}^{M\times N}$, we rasterized the $S^2$ cells from $\text{base}(M,f^*)$ to the common image following Algo.~\ref{algo:geoest} (see Sec. \ref{sec:ensembling_algo}).


The LS and LC TIFFs were downsampled by factor of 4 and 12 respectively, to the common image resolution. The binary masks in Algo.~\ref{algo:model_isec} were obtained from the downsampled images. Due to sparse sampling, some LC classes were merged by visual similarity prior to ensembling. Specifically, LC classes including cropland were merged (values 10-40 in \cite{esa2017}), as were broad-leaf tree cover (50--60), and the remaining tree cover on dry land (70--100). Short vegetation, including lichen and mosses (110--150), and vegetated and flooded regions (160--190) were also merged.  Urban areas were not merged, while bare areas were merged with water and permanent snow and ice (200--220). As a result, our models predict 7 classes for LC after merging.\label{sec:lc-merge}


\begin{table*}
      \centering

\footnotesize
\begin{minipage}{\textwidth}
        \centering

  \begin{tabular}{p{2.2cm}|p{0.6cm}|p{0.7cm}|p{0.7cm}|p{0.8cm}|p{1cm}||p{0.6cm}|p{0.7cm}|p{0.7cm}|p{0.8cm}|p{1cm}}
    \toprule
& \multicolumn{5}{c||}{RvA} & \multicolumn{5}{c}{GCD} \\
    \midrule

  &    {\bf Street} &     {\bf City} &    {\bf Region} &    {\bf Country} &  {\bf Continent} 
  &    {\bf Street} &     {\bf City} &    {\bf Region} &    {\bf Country} &  {\bf Continent} \\

 {\bf Method}
  &    {\bf 1~km} &     {\bf 25~km} &    {\bf 200~km} &    {\bf 750~km} &  {\bf  2500~km} 
  &    {\bf 1~km} &     {\bf 25~km} &    {\bf 200~km} &    {\bf 750~km} &  {\bf  2500~km} \\

\midrule
 GeoEst.
 &  --- & {\bf 39.42} & {\bf 65.98} & 85.37 & 97.02
 & {\bf 4.21} & {\bf 23.24} & {\bf 31.25} & 43.89 & 60.97 \\

 +LS
 & ---  & 35.08 & 65.58 & 85.89 & {\bf 97.45} 
 & 3.33 & 20.79 & 28.94 & 41.02 & 58.94 \\
 
 +LS, urban prior
 & ---  & 31.60 & 43.20 & 47.45 & 47.50 
 & 3.61 & 21.25 & 28.94 & 42.22 & 59.77 
 \\
 +LC
 & ---  & 36.91 & 65.90 & {\bf 86.78} & 97.23
 & 3.43 & 22.55 & 31.20 & {\bf 43.94} & {\bf 61.02} \\

 +LS+LC
 & ---  & 32.75 & 61.81 & 84.20 & 97.07 
 & 3.01 & 20.46 & 28.70 & 41.16 & 58.94 \\


  
\midrule
GeoCLIP
&  --- & {\bf 41.19} & 68.34 & 88.42 & 97.40
&  {\bf 4.91} & {\bf 30.37} & {\bf 43.84} & 62.22 & {\bf 79.81} \\

 +LS
 &  --- & 40.96 & {\bf 70.97} & {\bf 90.57} & 97.84
 &  4.07 & 28.61 & 42.59 & 61.39 & 78.98 \\

 +LS, urban prior
 &  ---  & 32.96 & 44.18 & 47.22 & 47.22
 &  4.35 & 27.96 & 41.34 & 61.34 & 78.94 \\

 +LC
 &  ---  & 40.29 & 71.29 & 90.24 & {\bf 97.86} 
 &  4.40 & 29.12 & 43.01 & {\bf 62.82} & 79.68 \\
 
 +LS+LC
 &  ---  & 37.48 & 69.14 & 89.88 & {\bf 97.86} 
 &  3.80 & 27.82 & 42.31 & 61.30 & 78.75  \\

\bottomrule
  \end{tabular}
\vspace{-2ex}
\caption{Results on Im2GPS3k\cite{vo2017revisiting} (balanced)} 
\label{table:results_im2gps}
\vspace{-3mm}
\end{minipage}

\end{table*}

\begin{table*}[ht!]
      \centering

\footnotesize
\begin{minipage}{\textwidth}
        \centering

\begin{tabular}{p{2.2cm}|p{0.6cm}|p{0.7cm}|p{0.7cm}|p{0.8cm}|p{1cm}||p{0.6cm}|p{0.7cm}|p{0.7cm}|p{0.8cm}|p{1cm}}
    \toprule
& \multicolumn{5}{c||}{RvA} & \multicolumn{5}{c}{GCD} \\
    \midrule

  &    {\bf Street} &     {\bf City} &    {\bf Region} &    {\bf Country} &  {\bf Continent} 
  &    {\bf Street} &     {\bf City} &    {\bf Region} &    {\bf Country} &  {\bf Continent} \\

 {\bf Method}
  &    {\bf 1~km} &     {\bf 25~km} &    {\bf 200~km} &    {\bf 750~km} &  {\bf  2500~km} 
  &    {\bf 1~km} &     {\bf 25~km} &    {\bf 200~km} &    {\bf 750~km} &  {\bf  2500~km} \\


\midrule
 GeoEst.
 & --- & 7.96        & 34.73       & 68.21     & 92.45
 & 0.00 &  5.00  & 11.03 & 33.53 & 60.44 \\

 +LS
 & --- & {\bf 9.85}  & {\bf 42.14} & 74.27     & 94.39 
 & {\bf 0.15} &  5.59  & 13.82  & 36.18 & {\bf 62.35} \\
 
 +LS, urban prior
 & --- & 9.36        & 33.63       & 41.06     & 41.21 
 & 0.00 & 5.00 & 11.18 &  34.12 & 60.88 \\
 
 +LC
 & --- & 8.29        & 40.25       & {\bf 75.43} & 94.37 
 & 0.00 & 5.15 & 11.47 & 35.29 & 61.32 \\
 
 +LS+LC
 & --- & 9.39        & 39.83       & 73.95     &   {\bf 94.67} 
 & {\bf 0.15} & {\bf 5.74} & {\bf 14.12} & {\bf 36.76} & 62.06 \\

\midrule


GeoCLIP
& ---  & 13.28       & 44.70       & 77.51       & 93.97 
& 0.00 & {\bf 12.43} & 29.24 & {\bf 64.18} & 81.14 \\

 +LS
 &  --- & {\bf 16.56} & {\bf 54.42} & {\bf 83.37} & 94.62 
 &  {\bf 0.44} & 12.28 & {\bf 30.26} & {\bf 64.18} & {\bf 83.33} \\
 
 +LS, urban prior
 &  --- & 15.76       & 38.06       & 42.03       & 42.04 
 &  0.00 & {\bf 12.43} & 29.39 & 63.89 & 81.58 \\
 
 +LC 
 &  --- & 15.69       & 51.33       & 82.69       & 94.97 
 &  0.00 & {\bf 12.43} & 30.12 & 63.45 & 80.70 \\
 
 +LS+LC 
 & ---  & 16.09       & 53.25       & 83.21       & {\bf 95.03} 
 & {\bf 0.44} & 12.28 & 30.70 & 63.60 & 83.19 \\

\bottomrule

\end{tabular}\vspace{-1ex}
\end{minipage}

\caption{Results on Street View\cite{luo2022g} (balanced)}
\label{table:results_streetview}
\vspace{-3mm}
\end{table*}

\subsection{Results}\label{sec:results}

We evaluate the geolocation methods using Algo.~\ref{algo:eval} for both Im2GPS3k and the Street View dataset, based on pano-IDs from \cite{luo2022g}. 
Results are presented in Figs.~\ref{fig:geoest_recall_vs_area} and \ref{fig:geoclip_recall_vs_area},  and Tables~\ref{table:results_im2gps} and \ref{table:results_streetview}.

Both Im2GPS3k and Street View datasets have uneven population density distributions (Im2GPS3k is predominantly urban, while Street View has relatively more non-urban).  To reduce this bias and better assess the effects on each population density type, we created rebalanced versions of the datasets by randomly sampling an equal number of images from each LandScan mask bucket.  We evaluate on both the original and rebalanced distributions.

The results are summarized in terms of recall vs. area (RvA) in Figs.~\ref{fig:geoest_recall_vs_area} and \ref{fig:geoclip_recall_vs_area}, both with and without balancing, and Tables~\ref{table:results_im2gps} and \ref{table:results_streetview} present results for the balanced imagery. We show results for the baseline rasterized algorithm for GeoEstimation or GeoCLIP predictions (Algo.~\ref{algo:geoest}); ensembling (Algo.~\ref{algo:model_isec}) with GeoEstimation or GeoCLIP combined with LS (+LS); GeoEstimation or GeoCLIP combined with LC (+LC); and GeoEstimation or GeoCLIP combined with both LS and LC (+LS+LC). 
Black dashed lines in the figures indicate areas of spherical caps with radii corresponding to the GCD thresholds typically used in literature, e.g., \cite{muller2018geolocation,cepeda2023geoclip}. It should be noted that the resolution of our common image prevents locating images within the 1~km GCD threshold (see Sec.~\ref{sec:ensembling}). The recall of the remaining ensembles is summarized as a function of global spherical cap areas corresponding to typical GCD thresholds used in literature with radius of the thresholds ($3.1$, $2.0\times10^3$, $1.3\times10^5$, $1.8\times10^6$, and $2.0\times10^7\text{km}^2$). Since in our proposed approach we accumulate all proposed areas in the final probability map (per Eq.~\ref{eq:recall_fcn_area}), recall is considered as a function of the total area of these often discontinuous regions.

We also quantify performance using conventional GCD. For direct comparison, the top-1 accuracy is computed using the common grid we used to implement RvA. The prediction is chosen as the highest-probability location from the common grid; note that due to the $\sim 4$~km grid resolution and the fact that pixels in non-contiguous areas can share values, there is no clear analog to a ``cell center'' in our approach, so this selection method results in somewhat worse GCD relative to published results.

Because GCD only evaluates the top-1 location, it is not able to capture the gains our method provides in limiting region extent: indeed, our method shows little improvement in the GCD for the single most confident point, even though it produces tangible gains in performance, as shown by the RvA metric.



For Im2GPS3k, the image counts for each LandScan mask (from lowest to highest population density) are $[540, 548, 1159, 715]$.
To rebalance, we randomly sample 540 images from each bucket, resulting in a total of 2160 images. Similarly, for Street View, the images in each mask bucket are $[11846, 17623, 18668, 1692]$, with the smallest number of images in the highest population density mask. Rebalancing results in 6768 images.


In Figs.~\ref{fig:geoest_recall_vs_area}, we observe larger gains for Street View, with the most significant gains occurring at the larger area thresholds, which generally correspond to rural areas. Note that there is also a performance drop for both base models on the out-of-domain Street View dataset compared to the typical Im2GPS3k evaluation set; however, we are able to recover a portion of this difference. Figs.~\ref{fig:geoest_recall_vs_area}--\ref{fig:geoclip_recall_vs_area}(c) display the absolute improvement in recall over the base models.

We also include a constant ``urban prior'' model as an additional baseline, labeled ``+LS, urban'', which always applies a mask restricting the data to the two highest-density LS buckets. Since the original imagery is primarily drawn from urban regions, such a constant prior might be effective. However, we find that the +LS attribute predictors outperform the constant prior, as expected, particularly in larger area regions. The urban prior model discards these regions wholesale, whereas predicted population density improves performance for these areas. Thus, the gains are not simply due to applying an urban region mask, but rather predicting different density attribute values is necessary to achieve improvements.

For the rebalanced datasets, we observe overall reductions in recall for the Im2GPS3k set and improvements for Street View. This is because the original Im2GPS3k dataset is biased toward regions with the highest population density, whereas Street View samples from the highest density areas considerably less. 
Using our ensembling approach, recall on Im2GPS3k is reduced at smaller area scales for both the original and rebalanced datasets, but improvements are seen at larger scales (see Fig.~\ref{fig:geoest_recall_vs_area}--\ref{fig:geoclip_recall_vs_area}, Tables~\ref{table:results_im2gps} and \ref{table:results_streetview}). Peak improvements in absolute recall are 0.21 for GeoEst and 2.96 for GeoCLIP base models.


At smaller scales, e.g. spherical cap areas corresponding to 25 and 200~km, the +LS ensemble gave the largest improvements.  The +LC and full +LS+LC ensembles provide benefits at larger scales.




The benefits of ensembling the ground-level attribute predictors at a global scale with the baseline image geolocation predictors are clear. The attribute predictors bring additional ground-truth information to the solution and improve performance relative to baseline.

\section{Conclusions and Future Work}

We have introduced a new image geolocation metric, {\it Recall vs Area} (RvA), along with an ensembling approach to global-scale image geolocation that incorporates ground-level attribute predictors based on satellite data products. Our metric addresses the challenge of accounting for ground-level image similarity across different geographic regions and can measure discontiguous predicted areas. Since practical image geolocation systems often prioritize search area, RvA naturally aligns with these needs. By ensembling our ground-level attribute predictors, we improved upon the state-of-the-art (SOTA) image geolocation method, GeoCLIP \cite{cepeda2023geoclip}, in terms of RvA.

Our ensembling approach enhances image geolocation recall as a function of search area by combining ground-level attribute predictors with geolocation models. These attribute predictors leverage satellite-based attribute masks, adding valuable information to the solution and improving performance over strong baselines, particularly in regions under-sampled in the original dataset.

Interestingly, the relative gain from ensembling predicted attribute masks was greater for GeoCLIP than for GeoEst, even though GeoCLIP began with a higher performance level. This suggests that the embeddings used in GeoCLIP may be complementary to our ground-level attributes, an area we hope to explore in future work.

By linking satellite-recognizable attributes to ground-level images, we can extend location information to areas of the world not represented in geotagged image data but included in satellite coverage. We have validated this approach using population density and land cover, and aim to extend it to additional attributes. 



{\small
\bibliographystyle{ieee_fullname}
\bibliography{refs}
}



  

\end{document}